\documentclass[11pt]{article}
\usepackage{endnotes}
\usepackage{enumerate}
\makeatletter
\renewcommand*\makeenmark{\hbox{\textsuperscript{[\@Alph{\theenmark}]}}}
\makeatother

\usepackage{hyperref}
\usepackage{times}
\usepackage[]{natbib} 
\usepackage{amsmath}
\usepackage{amssymb}
\usepackage{graphicx}
\usepackage[usenames,dvipsnames,svgnames,table]{xcolor}
\usepackage[T1]{fontenc}
\usepackage[utf8]{inputenc}

\usepackage{epigraph} 
\makeatletter
\def\@seccntformat#1{\@ifundefined{#1@cntformat}%
   {\csname the#1\endcsname\quad}  
   {\csname #1@cntformat\endcsname}
}
\let\oldappendix\appendix 
\renewcommand\appendix{%
    \oldappendix
    \newcommand{\section@cntformat}{\appendixname~\thesection\quad}
}
\makeatother

\begin{document}

\title{Verbal behavior without syntactic structures:\\ beyond Skinner and Chomsky}



\author{
  Shimon Edelman \\ Department of Psychology \\ Cornell University \\
  \href{https://shimon-edelman.github.io}{https://shimon-edelman.github.io}
}

\date{February 4, 2019}

\maketitle

\begin{abstract}
  
  What does it mean to know language? Since the Chomskian revolution,
  one popular answer to this question has been: to possess a
  generative grammar that exclusively licenses certain syntactic
  structures. Decades later, not even an approximation to such a
  grammar, for any language, has been formulated; the idea that
  grammar is universal and innately specified has proved barren; and
  attempts to show how it could be learned from experience invariably
  come up short. To move on from this impasse, we must rediscover the
  extent to which language is like any other human behavior: dynamic,
  social, multimodal, patterned, and purposive, its purpose being to
  promote desirable actions (or thoughts) in others and self. Recent
  psychological, computational, neurobiological, and evolutionary
  insights into the shaping and structure of behavior may then point
  us toward a new, viable account of language.

\end{abstract}

\begin{epigraphs} 
  \qitem{%
      \textit{Requiescat} \\
      \ \\
      Jade bowl \\
      Of the greatest antiquity \\
      ’Bout time it got broke. \\
      \ 
  }{--- Warren S.\ McCulloch \citeyearpar{McCulloch65freud}}
\end{epigraphs}

\section{The Chomsky Option}
\label{sec:one}

If proposing a theory that precipitates a paradigm shift in several
disciplines and becomes a byword for originality is a sign of a
scientist's influence, then Chomsky's legacy is secure. The theory
does not have to be right. This last point is central to Dennett's
Deal --- a Faustian dilemma, which serves as a commentary on varieties
of scientific temperament, ambition, and achievement.\footnote{For the
  interview in which Dennett discusses those matters, see
  \href{https://www.edge.org/conversation/dennetts-deal}{https://www.edge.org/conversation/dennetts-deal}.}
Chomskian linguistics makes a cameo appearance in the vicinity of the
dilemma's second horn:

\begin{quote}

  If Mephistopheles offered you the following two options, which would
  you choose?
  
  (1) to win the race (and the accompanying Nobel Prize!) for pinning
  down a discovery that became the basis for a huge expansion of
  scientific knowledge [\dots];

  (2) to propose a theory so original, so utterly unimagined before
  your work, that your surname enters the language --- but your theory
  turns out to be dead wrong, though it continues to generate
  centuries of arguably valuable controversy (I think of Lamarckian
  theories of evolution, and Cartesian theories of the mind. The jury
  is still out on Chomskian linguistics. It certainly passes the
  originality test. Like the victory of the America in the race that
  gave the America's Cup its name, there was no second anywhere in
  sight when Chomsky burst on the scene.)

  \begin{flushright}
    --- D.~C.~Dennett (1999)
  \end{flushright}

\end{quote}

\noindent
In the couple of decades that went by since that interview with
Dennett, the case against Chomskian linguistics gained much
momentum.\footnote{A capsule review of the current version of
  Chomsky's theory appears in \citep{EveraertEtAl15}. The opposition
  to it comes from many quarters, represented by
  \citep{Tomasello98npl,Lamb04,Postal04,EvansLevinson09,RamscarPort16}.}
It seems to me, however, that Chomsky's accomplishment extends beyond
showcasing option (2) in Dennett's Deal. Independently of whether his
theory of language is right or ``dead wrong,'' Chomsky's contribution
exemplifies --- indeed, epitomizes --- a third option, not broached by
Dennett:

\begin{quote}

  (3) to define the parameters of the discourse, in an entire cluster
  of disciplines, so decisively that even people who oppose your
  theory are forever forced to argue on your terms.

\end{quote}

\noindent
The discourse parameters set by Chomsky in an extensive body of
extremely influential work, beginning with his \textit{Syntactic
  Structures} \citeyearpar{Chomsky57} and review of B.~F.~Skinner's
\textit{Verbal behavior} \citeyearpar{Chomsky59}, can be succinctly
characterized by a tripartite dogma, stated next.

\subsection{The grammar dogma}
  
The following three interlinked assumptions are nearly universally
accepted by linguists, regardless of their theoretical persuasion:

\begin{enumerate}
\item that the \emph{sentence} is the unit of language production and
  comprehension;
\item that some sentences are \emph{well-formed}, while others are not;
\item that sentences are generated by a set of formal syntactic rules
  --- a \emph{grammar} --- enforcing well-formedness in production and
  relying on it in comprehension.
\end{enumerate}

\noindent
The concept of grammar, which anchors the Chomskian dogma, has proved
to be particularly pervasive and has spread far beyond the boundaries
of linguistics. It has been and is being invoked by theorists in
domains of inquiry ranging from psychology, through animal
communication, music, anthropology, and various branches of philosophy
(of mind; of language; of psychology; of science; of morals), to
literary theory, history, archaeology, and cultural
analysis.\footnote{I have relegated the relevant references to
  Endnote~[A] to avoid distracting the reader
  inordinately.}\endnote{\label{e:1} What follows is a very partial
  list of references to the long reach of the concept of grammar,
  ordered by publication year and focusing, where possible, on
  critical reviews:
  \begin{itemize}
    \itemsep 0pt
    \item \emph{Psychology} --- according to \citet{Miller03}, the
      roots of the ``cognitive revolution'' can be traced back to his
      reading of Chomsky's \textit{Syntactic Structures}
      \citeyearpar{Chomsky57}.
    \item \emph{Music} --- for instance, \textit{In search of a
      generative grammar for music} \citep{Laske73}.
    \item \emph{Literary theory} --- for instance, \textit{Linguistic
      models and recent criticism: transformational-generative grammar
      as literary metaphor} \citep{Henkel90}.
    \item \emph{History} --- ``In a recent chapter on the
      post-Westphalian development of sovereign states, Krasner denies
      that a deep generative grammar shaped European state formation
      [\dots]'' \citep[p.121]{ReusSmit99}.
    \item \emph{Archaeology} --- for instance, \citep{HodderHutson03}.
    \item \emph{Anthropology} --- ``\textit{Every culture is generated
      within a certain frame; cultures are the concrete actualizations
      of civilization, which they turn into an applied worldview} -- a
      bit like the transformations in generative grammar''
      \citep[p.30]{Francois07}. 
    \item \emph{Philosophy and psychology of morality} --- for
      instance, \textit{Universal moral grammar: a critical appraisal}
      \citep{JacobDupoux07}.
    \item \emph{Cultural analysis} --- for instance, \citep{Born10}.
    \item \emph{Philosophy of science} --- as in the concept of
      ``Galilean linguistics'' \citep{Kertesz12,Behme13}.
    \item \emph{Ethology} --- here is a typical argument in favor of
      adopting Chomsky's framework in the study of monkey
      vocalizations: ``We would argue that even in the case of human
      language, where the existence of sophisticated cognitive
      abilities is uncontroversial, a precise characterization of the
      formal properties of a language is a good first step towards an
      analysis of its cognitive implementation. In syntax, various
      approaches broadly inspired by Chomsky’s work (e.g.\ 1957, 1965)
      developed a formal and a cognitive approach in tandem''
      \citep[p.13]{SchlenkerEtAl16}.
  \end{itemize}
}

In linguistics and in the psychology of language, it is nearly
impossible these days to formulate, test, and defend a theory that
does not take a stance with regard to the nature and role of
grammar.\footnote{This can only be expected, given how polemical
  discourse typically works \citep{ApothelozEtAl93}.} Even researchers
who would rather stay clear of the concept of grammar find themselves
arguing against it,\footnote{Not excepting the present chapter.}
thereby lending a certain legitimacy to a conceptual framework which,
should it go down, would take along linguistics as we know it,
re-revolutionize cognitive psychology,\footnote{Thereby overturning
  the regime established after the first cognitive revolution
  \citep{Miller03}.}  and send aftershocks through many parts of the
human intellectual endeavor.

Given the scope of the problem of grammar, and how it affects both the
formalist, Chomskian orthodoxy and the various ``heresies''
(functionalist, usage-based, etc.), it is fascinating to observe how
much of it is due to poor meta-theoretical choices. These, in turn,
signify a disregard of what is otherwise considered to be the standard
methodology in the cognitive sciences.

\subsection{On the proper methodology in behavioral and brain science}

What computational structures and algorithms give rise to linguistic
behavior? At present, most linguists --- including those who hold that
humans (and only humans) possess a language organ
\citep{HauserChomskyFitch02} and who therefore claim to practice
``biolinguistics'' \citep{Fitch09} --- put this
computational/algorithmic question first:

\begin{quote}
  We need to distill what we know from linguistic theory into a set of
  computational primitives, and try to link them with models and
  specific principles of neural computation. [\dots] We need a list of
  computations that linguistic theorists deem indispensable to solve
  their particular problem (e.g., in phonology, syntax, or semantics)
  \citep[p.298]{Fitch09}.
\end{quote}

\noindent
Now, building a research program on ``computations that linguistic
theorists deem indispensable to solve their particular problem'' can
only work if the linguists have got the \emph{problem} right. Therein
lies the rub: across all cognitive domains, it is this abstract,
problem level (using Marr's \citeyearpar{Marr82} terminology) that
presents the greatest, yet least appreciated difficulties. On this
level, it is all too easy to fall in with an intuitively appealing
hypothesis, which may then lead the entire research program down a
dead-end path. A case in point is vision, where, following Marr's
\citeyearpar{Marr82} own influential arguments, it was believed for a
long time that the overarching problem is to reconstruct the
three-dimensional layout of the visual world and the volumetric part
structure of objects. This theoretical approach ultimately proved to
be barren
\citep{BarrowTenenbaum93,Dickinson97panel,Edelman02tics,Edelman09tosee},
and has been gradually abandoned in favor of the use of learning to
map visual data to various task-dictated representations
\citep{Edelman12marr,KrizhevskySutskeverHinton12,Oliva13}.

For this reason, the current consensus view in cognitive sciences,
forty years after Marr (\citeyear{MarrPoggio77}; see also
\citealp{Marr82,Poggio12}), is that representation- and
algorithm-level questions cannot be settled without also addressing
the other, complementary levels of understanding: the problem level
(what is it, in terms of computation, that needs to be done, and why?)
and the implementation level (how are the posited algorithms realized
by the system?).  Moreover, for any natural biological computational
phenomenon such as language, the Marr hierarchy must be extended to
include evolution, development, and behavioral ecology
\citep{Mayr61,Tinbergen63}, which are all interrelated
\citep{LalandEtAl11}.

Any research program in linguistics must, therefore, be examined for
tacitly assumed, and possibly wrong, answers to questions that arise
on various levels of the Mayr/Tinbergen/Marr hierarchy.\footnote{A
  mistaken view, on which the levels are independent and can be
  studied in isolation, appears to be prevalent in linguistics. Thus,
  while \citet{WalenskiUllman05} refer to the importance of levels of
  explanation as ``another tenet of the cognitive revolution,'' they
  proceed to state that ``On this view, a given phenomenon is best
  accounted for at a particular level of explanation that is largely
  independent from other levels.''} In this paper, I do so while
focusing on programs that belong to two camps: formalist, whose
adherents hold that the core issues in linguistics have to do with the
structure of utterances, and functionalist, where the overarching
concern is with the use of language in communication \citep[for a
  discussion of this distinction, see, e.g.,][]{Newmeyer16}.  The
history of modern linguistics is the history of the tension between
these two camps, which stems from the disparity between the answers
they give to the fundamental problem-level question:

\begin{itemize}
\item What is language for? (Section~\ref{sec:for})
\end{itemize}

\noindent
It may be surprising to some that the answers offered by formalists
and by functionalists to the next two questions

\begin{itemize}
\item What is language like? (Section~\ref{sec:like})
\item What does it mean to know language? (Section~\ref{sec:grammar})
\end{itemize}

\noindent
are identical; both hinge on the concept of grammar. To decide whether
or not this has been a good methodological choice (that is, whether
the assumption it incorporates was ultimately fruitful), I then ask:

\begin{itemize}
\item Where has this gotten us? (Section~\ref{sec:where})
\end{itemize}

\noindent
The conclusions of the ensuing review of the achievements of the
formalist and functionalist research programs are rather
bleak,\footnote{Some of the points I make here merit book-length
  treatment. In formulating them, I opted for brevity: rather than
  pretending that the issues can be easily settled, my goal is to
  provoke doubt, with regard to even the most hallowed assumptions of
  the orthodoxy.} which is all the more reason to move on to the last
question:

\begin{itemize}
\item Where should we go from here? (Section~\ref{sec:next})
\end{itemize}

\noindent
In that section, I outline an alternative research program that aims
to engage with all the levels of the Mayr/Tinbergen/Marr hierarchy:
evolution, development, behavior, computational problems, algorithms,
and implementation in the brain.

\section{What is language for?}
\label{sec:for}

Asking this kind of question about a mental faculty\footnote{Or even
  about an aspect of an animal's anatomy or physiology.} is always a
useful exercise, because the intuitively obvious answer so often ends
up being wrong. What is memory for? The storehouse metaphor, which had
been a staple of psychology textbooks ever since the cognitive
revolution, and which suggests that memory is ``for storing
information,'' has by now been largely rejected
\citep{KoriatGoldsmith96,Glenberg97bbs,GilbertWilson07}. Indeed, the
very notion that memory involves a definite number of discrete storage
slots (as per G.~A.~Miller's \citeyearpar{Miller56} famous title
``magical number seven, plus or minus two'') is now seen as a baseless
idealization \citep{MaHusainBays14,BradyEtAl16}. Instead, a plausible
concise answer to the memory question at present is that memory works
``in the service of perception and action'' \citep{Glenberg97bbs} ---
in other words, memory is for behavior.

A similar process of transcending the obvious, and wrong, answer is
under way in vision. What is vision for? The opening passage of Marr's
\textit{Vision} \citeyearpar{Marr82}, one of the most influential
books in cognitive science, stated that ``the plain man's answer (and
Aristotle's, too) would be, to know what is where by looking.''
Although this idea still drives much of the research in ``neural
networks''-style AI, it is increasingly recognized as inadequate in
robotics \citep{AloimonosWeissBandopadhay88,Ballard91,BreazealEtAl01}
and especially in theories of biological vision
\citep{Noe04,Edelman05noe,Edelman12marr,Edelman15}. It seems that
vision is \emph{for} the same thing memory is for: behavior
\citep{Edelman16ox}.  What, then, could language be for?

\subsection{The formalist approach}
\label{sec:Minimalist}

Relying too much on common sense in deciding what language is for is
not something that formalist linguistics can be accused of.  Marr's
``plain man'' (and perhaps Aristotle, too) would likely be surprised
by the Chomskian answer to this question, namely, that language is
\emph{for} structured thinking \citep{EveraertEtAl15}.  Isn't language
the ultimate tool of human sociality, which has evolved in the service
of, and used for, communication? Not according to the Minimalist view,
which holds that the use of language in communication is derivative
and ``ancillary'' \citep{EveraertEtAl15} to its use as a formal
computational system, or grammar, for structuring thoughts. On this
view, communication is mediated by two interfaces: one that maps back
and forth between the grammar and logical forms that express meanings
and the other that maps between the grammar and sequences of motor
acts in the auditory or gestural medium. Through this second
interface, the grammar of a language supposedly sanctions all and only
the well-formed sentences (that is, stand-alone utterances) in that
language.

\subsection{The functionalist approach}

An alternative take, which justifies my choice of the
``functionalist'' label for the theories that share it, is that
language is \emph{for} communication, because that's what the overt
function of language seems to be.  While language is clearly extremely
useful in assisting and augmenting human thinking
(A.~\citealp{Clark98}), most psychologists (and some linguists) hold
that the primary utility of language and the origin of its versatility
are in the social domain: ``Language is first and foremost an
instrument of communication --- the ``\textit{exchange} of thoughts,''
as one dictionary puts it --- and it is only derivatively an
instrument of thought'' (H.~\citealp[p.325]{Clark96}; see also
\citealp{Bruner75,BatesMacWhinney82,PinkerJackendoff05}). In this
framework too, thoughts that are to be communicated take the form of
well-formed sentences, each with a well-defined meaning, as per the
standard view of what communication is and how it works \citep[for a
  recent, computationally sophisticated and explicit example,
  see][]{GoodmanFrank16}.

\section{What is language like?}
\label{sec:like}

The shared focus on sentences and well-formedness characterizes also
the respective answers offered by formalists and functionalists
regarding the question of what language is like. Indeed, here there is
even less of a difference between the two frameworks.

\subsection{The formalist approach}

While the details of the formalist answer to the question of what
language is like have changed more than once between the publication
of \textit{Syntactic Structures} \citep{Chomsky57} and \textit{The
  Minimalist Program} \citep{Chomsky95}, the core story has remained
the same. It goes roughly as follows. Although language superficially
appears to consist of linear sequences of gestures (vocal, visual, or
others), what it \emph{really} consists of, and what determines
whether or not sentences (standalone, self-contained sequences) are
well-formed, is their hidden hierarchical structure --- specifically,
labeled trees in which both intermediate nodes and leaves may belong
to abstract categories, including empty elements and traces left by
elements that moved elsewhere.\footnote{For a typical example, plucked
  at random from a huge literature, see \citep[Fig.1]{Graf12}.}

The notion of syntax \citep{Shibatani15} within this framework
corresponds to a set of phenomena that are derived from two key
assumptions: (i) the discrete infinity of language and (ii) the
existence in utterances of structure that transcends the mere linear
sequencing of elements \citep{Phillips03encyc}. (Although these
assumptions are treated by syntacticians as self-evident axioms, they
can and have been questioned, as we'll see in Section~\ref{sec:next}.)
A corpus of language is thus supposed to consist of well-formed
sequences of potentially unbounded length that possess hidden
structure, characterized by constituency, hierarchy (c-command,
scope), various types of dependencies, and constraints on
dependencies.\footnote{The same conceptual domain of syntactic
  phenomena is sketched by \citet[ch.3]{TownsendBever01}, under the
  heading ``What every psychologist should know about grammar.''
}

\subsection{The functionalist approach}

The functionalist answer to the question of what language consists of
is identical to the formalist one: hierarchical structures over
discrete elements, which may include abstract category labels. For
instance, an early construction grammar analysis of the sentence
``What did Lisa buy the child?''  includes, in addition to the six
trivial, fully lexicalized constructions (words), five fully abstract
ones: ditransitive, question, subject-auxiliary inversion, VP, and NP
\citep[fig.1]{Goldberg03}.



In another example, \citet[Figures 3-6]{ArbibLee08} describe the
hypothetical process of mapping a meaning to a sentence that expresses
it as beginning with a semantically annotated tree structure that
encodes the desired meaning, and proceeding through a series of
derivations to cover that tree with another tree --- a kind of phrase
structure representation from which the resulting sentence can be read
out.\footnote{Note that unless the sentence --- the carrier of meaning
  --- is complete and well-formed, it cannot be formally interpreted:
  the syntactic parser would crash before the semantic interpretation
  stage can even be reached.}

\subsection{The language myth}

The above comparison reveals that the difference between formalists
and functionalists regarding the nature of language is rather
superficial. On the formalist account, which puts thinking first, a
\emph{thought} is a well-formed sentence, to which there is attached a
meaning. In comparison, on the functionalist account, which focuses on
communication, a \emph{message} is a well-formed sentence, to which
there is attached a meaning. Thus, in both cases, grammar rules the
day, so to speak, insofar as it underlies sentence well-formedness.


This preoccupation with grammar and syntactic well-formedness
handicaps any theory of semantics that accedes to playing second
fiddle to syntax \citep{Lakoff71,Partee14}. Merely elevating semantics
above syntax is, however, of no avail, as long as one does not
acknowledge an even more serious theoretical liability: the view of
language as a code. As \citet[p.529]{Love04} states it,

\begin{quote}
  It is the idea that linguistic \emph{communication} takes place in
  virtue of shared knowledge of the particular inventory of
  form-meaning correlations that constitutes the language in
  use. Communicating is a matter of encoding and decoding messages in
  terms of the shared code. In fact, these two ideas form a closed
  circle: the fixed-code concept of a language explains how
  communication is possible, while if this is what communicating is,
  then languages have to be fixed codes to enable communication to
  take place.  This, in a nutshell, is what Roy Harris calls the
  `language myth' \citep[see e.g.][]{Harris81}.
\end{quote}

\noindent
Even if the communication of meanings (in the narrow sense of the
exchange of messages) is what language is for, it is, or should be, a
truism that the practical import of a message --- its effect on the
listener (and/or the speaker), which is a reasonable way of
operationalizing meaning --- is fully contained neither in the
sequence of words that comprise an utterance, nor in its combination
with a putative grammar that assigns to it some other, perhaps
hierarchical, structure \citep[see][]{RamscarBaayen13}. In
Section~\ref{sec:alt}, I sketch an alternative view of language, which
does not take it to be a code.

\section{What does it mean to know language?}
\label{sec:grammar}

At this point, it should be unsurprising that both formalist and
functionalist theories equate the knowledge of language with the
possession of grammar. Still, for symmetry and completeness' sake, I
shall proceed to state the obvious.

\subsection{The formalist approach}

The formalist accounts, with their postulates of discrete infinity and
hidden structure of language, are naturally centered on the notion of
grammar --- a formal system that generates all and only well-formed
sentences and is also used in parsing (recovering the structure of a
sentence). \citet{Stabler13b}, paraphrasing an expression from
\citep{Bever70}, called grammar ``the epicenter of linguistic
behavior.''  More recently, \citet{LauEtAl16} noted that ``[\dots] the
view of a formal grammar as a binary decision procedure has been
dominant in theoretical linguistics for the past 60 years.''  The
centrality of grammar and the presumed ``autonomy of syntax''
\citep[p.138]{Chomsky04}, are among the foundations of the most recent
incarnation of Chomsky's theory: the Minimalist Program
\citep{Chomsky95}, which holds that grammar --- a formal recursive
system for building hierarchical structures --- is in charge over the
other components of the language faculty: ``syntax creates structures
that are interpreted at the interfaces with the
Articularory-Perceptual [\textit{sic}] and Conceptual-Intentional
[meaning] systems'' \citep[p.141]{Irurtzun09}.

\subsection{The functionalist approach}

The functionalists may disagree among themselves as to what precise
form grammar takes, but they are nearly unanimous in endorsing the
notion of grammar itself: witness titles such as ``Functionalist
approaches to grammar'' \citep{BatesMacWhinney82}, ``Grammar and
conceptualization'' \citep{Langacker99}, ``Radical construction
grammar'' \citep{Croft01}, or ``Grammar as architecture for function''
\citep{DuBois03book}.  To the extent that modern functionalist
linguistic approaches arise out of opposition to, and are motivated by
the perceived failures of, formalist linguistics, one wonders whether
anything short of complete repudiation of the foundational tenets of
the latter would ever suffice. In what follows, I argue that it would
not.


\section{Where has this gotten us?}
\label{sec:where}

The short answer to this question, in my opinion, is: ``Not too damn
far.''\footnote{This view is not very common among linguists, but it
  is occasionally encountered. One variant is due to
  \citet[p.29]{Sampson07}: ``[\dots] Over the half-century during
  which the generative conception of grammar has been influential, it
  does not appear that the possibility in principle that grammars
  could tell us about minds has led to many specific discoveries about
  mental functioning or mental structure.'' Similarly,
  \citet[p.328]{WalenskiUllman05} state: ``What has \emph{not} emerged
  is a true science of language.''} Because the formalist and the
functionalist approaches hardly differ in their conceptual
foundations, in the brief critique of their achievements that I offer
in Section~\ref{sec:1} I lump them together. I then evaluate, in the
equally brief Section~\ref{sec:2}, a selection of empirical methods
that are light on linguistic theory, a category that includes most of
the work in applied natural language processing or NLP. As we shall
see, while some of the empirical successes have been very impressive,
they have little bearing on key issues in real natural language
learning and use.

\subsection{The formalist/functionalist approaches}
\label{sec:1}

My present case for the explanatory barrenness of grammar rests on two
sets of considerations: the nature of evidence required to
substantiate linguistic theories; and computational, psychological,
and neurobiological findings regarding the reality of syntactic
structures posited by these theories.

\subsubsection{On the required evidence in linguistics}

What kind of evidence should we require when evaluating a theory
involving grammar? If linguistics were merely a branch of mathematics
(as is formal language theory, to which Chomsky made seminal
contributions; \citealp{HopcroftUllman79}), the question of evidence
would have been moot: mathematicians are free to chose their axioms
and to use them to derive whatever theorems that follow. Grammarians
do, however, explicitly claim to be doing science, as indicated for
instance by Chomsky's long-standing insistence on calling his program
``Galilean linguistics'' (critiqued in detail by
\citealp{Behme14galilean}) and by the recent ascendance of
biolinguistics.  Thus, evidence \emph{is} required for claims about
the reality (or at least relevance) of posited syntactic structures,
over and above the possibility of deriving such structures from
``axioms'' (which would have sufficed in mathematics).

Grumbling about the scarcity of such evidence has been going on for as
long as formalist claims have been made. It tends to intensify
periodically, as linguistic theories are abandoned or mutate beyond
recognition.\footnote{E.g., \citet{TownsendBever01}: ``the rapid
  changes in syntactic theories have left psychologists in large part
  baffled as to how to integrate grammatical knowledge and behavior in
  rigorous models.''} Examples range from complaints by outsiders
about the apparent lack of empirical support for Minimalism
\citep{EdelmanChristiansen03} to a wholesale labeling by an eminent
linguist of the entire Chomskian enterprise as ``junk linguistics''
\citep{Postal04junk}.  The problem of evidence in linguistics has also
attracted the attention of philosophers of science (e.g.,
\citealp{Behme14}).

In the interests of brevity, I shall focus here exclusively on the
famous metatheoretical criterion due to \citet{Laplace12}: ``The
weight of evidence for an extraordinary claim must be proportioned to
its strangeness.'' Are the structures posited by grammar-based
theories of language extraordinary? The ``linguistic wars'' of the
1970s, which revealed major conceptual fault lines and left
linguistics fractured, were brought about by the perception on part of
a large contingent of linguists that syntactic complexity was getting
out of hand and out of touch with psychological reality: ``The
underlying structures of the Generative semanticists became more and
more abstract, often resembling first-order logic, and those
structures plus the syntactic rules to get from there to surface
structure often seemed `wild'. In one famous example, the underlying
structure for `Floyd broke the glass' had 8 clauses''
\citep[p.8]{Partee14}. 

Partee \citeyearpar{Partee14} proceeds to remark that ``What seemed
`wild' then [in early 1970s] might not now: the shocking number of
clauses (7 or 8) in Ross's deep structure for `Floyd broke the glass'
does not come close to the number of functional projections that now
intervene between various pairs of `familiar' syntactic categories in
respected generative analyses such as Cinque (1999).''  Linguists
working within the Minimalist framework routinely debate alternative
analyses which all tend to be that complex, without ever questioning
the reasonableness of this complexity. For instance, the syntactic
analysis of the sentence fragment ``A picture I like,'' discussed by
\citet[p.331]{Stabler01}, is a tree consisting of 30 nodes. Now, the
number of tree nodes between successive words has traditionally been
used as a measure of syntactic complexity. \citet{BrennanEtAl16}, who
use this measure in their search for brain activity evidence in
support of syntax (more about which below), consider a Minimalist
parse in which 18 nodes intervene between the words ``fallen'' and
``by'' in the sentence ``I wonder how many miles I have fallen by this
time.''

Even if we set aside the question of how plausible syntactic theory is
from the standpoint of neurobiology,\footnote{Questions of
  neurobiological plausibility arise, for instance, in connection with
  the claim that the brain harbors symbolic representations of dozens
  of ``functional categories'' posited by \citet{RizziCinque16} or
  that it implements the complex, logically constrained string
  rewriting \citep{Stabler13}, which is required for most types of
  syntactic derivations, notably the ones postulated by Minimalists,
  to work.} a strong sense of unease remains. When the syntactic tree
purporting to describe the structure of a four-word phrase does so in
terms of categories most of which have no lexical counterparts and is
too deep to fit on a journal page, something is out of joint. Whether
or not the claim of the reality of such structures is true, it comes
across, to me, as pretty extraordinary.

\subsubsection{On the available evidence in linguistics}

Are any of the historical or modern contenders for a theory of syntax
supported by evidence that is commensurate with the extraordinary
claims that are being made? It may seem reasonable to require that
linguists who propose to explain natural language by positing the
existence of a formal grammar that generates it take observable data
--- a corpus of real language --- and infer from it the grammar in
question. Unfortunately, from the standpoint of computational
complexity, formal language theory is a mixed bag: many interesting
problems, including grammatical inference for potentially relevant
classes of grammars, are intractable
\citep{ClarkLappin13,HeinzEtAl15}.  In any case, even when positive
results in learnability become available (e.g.,
\citealp{Clark13lang}), they are more likely to be argued away rather
than adopted by those syntacticians who follow Chomsky's ``argument
from the poverty of the stimulus'' (POS; \citealp{Chomsky86}) in
holding grammar to be unlearnable and therefore necessarily
essentially innate \citep{BerwickEtAl11}, as per the hypothesis of
Universal Grammar.\footnote{The POS argument (see
  \citealp{PullumScholz02} for a critical analysis) proposes that
  language must be innate because overt data never contain enough
  information for the proper generalizations to be learned from
  experience; a typical claim is that ``A person might go through much
  or all of his life without ever having been exposed'' to relevant
  information \citep[p.40]{PiatelliPalmarini80}. On the POS account,
  structural generalization is possible despite being unlearnable
  because it is supported by an innate ``Universal Grammar''; this
  notion is discussed and critiqued in references listed in the next
  paragraph.}

It should be noted that the denial of learnability of grammar
undermines the very theoretical enterprise that has engendered it: if
grammar is algorithmically unlearnable, how reliable can the
grammarians' intuitions about it be? The standard POS reply to this
concern is that ``positive data'' such as grammaticality judgments
that are relied upon by theorists (often to the exclusion of any other
data) are not available to developing infants. The POS stance does
not, however, withstand empirical scrutiny by fellow linguists
\citep{PullumScholz02,ClarkLappin11} and is generally rejected by
developmental psychologists and neuroscientists
\citep{Bates94,BatesGoodman99,Tomasello03book,GoldsteinEtAl10tics,EdelmanWaterfall07,SyalFinlay11}. The
same goes for Universal Grammar
\citep{Jacobson66,McCawley82,Comrie88,Bouchard12,EvansLevinson09,AmbridgePineLieven15}.

Be it as it may, none of the syntacticians' efforts so far has led to
the formulation of a sizeable-coverage (let alone complete) grammar
for any language.\footnote{While hand-constructed or automatically
  learned grammars may be good at parsing, they are typically very bad
  at generation. State of the art natural language generation systems
  \citep[e.g.,][]{SharmaEtAl16} shun grammar-based approaches (relying
  instead on ``neural networks'') and report human evaluations based
  on ``appropriateness'' instead of grammaticality; still, their
  performance, even in highly constrained domains, is middling at
  best.}  Instead, explanations are offered for selected syntactic
phenomena, which themselves are invariably formulated exclusively in
terms of abstract, and therefore not directly observable, constructs
of syntactic theory. When an account clashes with new ``data'' (that
is, binary judgments of grammaticality, typically produced by the
theorists themselves), the contradiction is ignored (following
Chomsky's principle of ``epistemological tolerance'';
\citealp{Kertesz12}), or else the assumptions underlying the original
analysis are modified or withdrawn in favor of new
ones.\footnote{``[\dots] How can anyone maintain the hypothesis of a
  universal grammar? The answer is to make the concept immune to
  falsification.''  --- M.~Tomasello, commenting on
  \citep{Everett05}.}  The theoretical discourse in the study of
grammar is thus markedly self-contained, with little or no input from
empirical methodologies\footnote{Even those rooted in formal language
  theory \citep[e.g.,][]{Kornai98}.}  or from neighboring disciplines.

Meanwhile, in the neighboring disciplines of psycholinguistics and
neurolinguistics (not to be confused with ``biolinguistics''),
attempts to glean relevant evidence have been under way all along. The
results do not bode well for grammar-based accounts. Two of the fronts
the news from which have been consistently bad are: (1) investigations
into the nature of grammaticality judgments and (2) studies that probe
the reality of various theoretical constructs advanced by grammarians.

That grammaticality (or ``grammaticalness''; \citealp{Chomsky57}) is
graded has been known to psycholinguists for a long time (e.g.,
\citealp{Kess76}),\footnote{And so has been also the dependency of
  grammaticality on context \citep{Langacker98} --- which shows that
  considering language one ``sentence'' at a time is a mistake.} and
has even been discussed by Chomsky \citeyearpar{Chomsky64}.  The full
extent of the problem that grammaticality judgments pose for formalist
linguistics became a major topic of discussion only relatively
recently \citep{Schutze96}.  Grammaticality judgments are the primary
source of data for students of grammar (a methodological choice that
depends entirely on the acceptance of the grammar
dogma). Nevertheless, the problems surrounding grammaticality have had
little effect on syntactic theorizing, which traditionally isolates
itself from psycholinguistics by claiming a distinction between
linguistic \emph{competence} --- hypothetically, perfect in all native
speakers, but unobserved --- and ``mere'' \emph{performance}, whose
imperfections mask the perfect competence, making it unobservable in
principle \citep{Chomsky65}. Given the immunity to falsification of
this distinction, in psycholinguistics the concept of grammaticality
has been all but displaced by acceptability. Nevertheless, the weight
of the accumulated evidence regarding its gradedness has attained such
proportions \citep{LauEtAl16} that even the impregnable
competence/performance wall is beginning to show cracks.\footnote{Yet,
  the wall still stands: \citet{LauEtAl16} make a point of noting that
  ``We are not of course denying the distinction between competence
  and performance.''}

Considering how central the existence of a brain basis for syntax is
to Chomskian (bio)linguistics, the scarcity of behavioral and brain
evidence for syntactic structures is striking. One relatively
uncontroversial set of findings has to do with the ``phrase
structure'' hypothesis, common to most theories of syntax, according
to which sentences possess hierarchical, in addition to linear,
structure. In an early study, \citet{Johnson65} showed that the error
rates of subjects who had memorized a set of multi-clause sentences
spiked at the boundaries between clauses posited by a structural
analysis and were higher, the higher in the syntactic tree the
boundary was supposed to reside. Half a century later, little has been
added to that finding: results obtained using state of the art
magnetoencephalography (MEG) brain imaging showed that, in subjects
listening to connected speech, ``cortical activity of different
timescales concurrently tracked the time course of abstract linguistic
structures at different hierarchical levels, such as words, phrases
and sentences'' \citep{DingPoeppel16}.

In comparison to the basic phrase structure, evidence supporting the
reality of more far-fetched theoretical constructs, such as movement,
traces/copies, etc., remains elusive.  The endemic problem is the
possibility of alternative explanations of response time or brain
activity differences: not in terms of the targeted syntactic
distinction, but rather in terms of some other distinction (pragmatic,
semantic, etc.) that had not been, and in many experimental designs
could not have been, controlled. It must be particularly disappointing
when a sophisticated syntactic account is overshadowed by a glaringly
primitive one, as seems to have happened in the impeccably conceived
and executed study of \citep{BrennanEtAl16}, mentioned earlier.

In that experiment, the time course of brain activity of subjects
listening to natural speech was regressed on several predictors,
including running estimates of syntactic complexity calculated from
several sources: node-counting approximations to Minimalist parses and
to parses generated by a context-free phrase structure grammar, as
well as Markov or $n$-gram language models with $n=2,3$. In the
regression, the magnitude of the $\beta$ coefficients of predictors
derived from Markov models and from the phrase structure grammar
exceeded those derived from the Minimalist grammar. While a step-wise
analysis suggested that the latter did make a distinct contribution,
it could easily have been due to some other neurocomputational
phenomenon that masqueraded as an increase in node count in the parse
tree. As with other attempts to prove that syntax is real (e.g., is
represented and processed as such in the brain), this pattern of
findings is not quite the extraordinary body of evidence that is
called for.

\subsection{Empirical work and NLP}
\label{sec:2}

The chronic lack of independent empirical support for theories of
syntax has prompted many engineers and (in cognitive psychology)
modelers interested in language to adopt a theory-neutral stance while
mounting algorithmic attacks on problems such as language acquisition,
parsing, and understanding. In natural language processing (NLP), this
approach is associated with the famous engineer's quip ``Whenever I
fire a linguist, our system performance improves'' \citep{Jelinek88}.
While NLP has made great progress in certain applied tasks (for
instance, achieving human parity on a natural speech transcription
benchmark; \citealp{XiongEtAl16}), all these tasks share the same
common characteristic: they reduce to mapping inputs to outputs, which
makes them amenable to supervised learning (particularly with state of
the art ``deep network'' technology, combined with massive amounts of
training data), yet mostly irrelevant to natural language behavior. As
I wrote elsewhere, deep networks ``can serve as a good model of
behavior only if we pretend that behavior amounts to one Google (or
Watson) query after another'' \citep{Edelman15}.

If it turned out to be possible to learn a decent-coverage grammar
from realistic data, the empirical approach would have justified
itself (in which case syntactic theory would at least have something
to look up to). At present, however, this possibility seems remote:
frontal attacks, algorithmic or heuristic, on language acquisition
either do not scale up, or result in precision/recall performance that
barely exceeds that of Markov models. Pretending that this is just
fine --- because the competition performs even worse, or perhaps
because the resulting model replicates certain features of human
performance --- exemplifies a kind ``soft bigotry of reduced
expectations,'' which characterized some of my own work on language
acquisition
\citep{SolanEtAl05pnas,KolodnyLotemEdelman15}.\footnote{This memorable
  phrase was coined by a speechwriter to the 43rd president of the
  United States. The idea that it may be applicable to the present
  case is not necessarily shared by my coauthors.}

This deplorable state of affairs is of course not specific to language
acquisition: it characterizes all linguistic endeavors that accept the
dogma, spelled out in Section~\ref{sec:one}, at the core of which are
the concepts of syntactic well-formedness and of grammar. For as long
as this dogma stands, all attempts to learn and use language, or to
understand how it works --- theory-neutral or theory-laden,
evidence-driven or ``Galilean'' --- will fail.

\section{Where should we go from here?}
\label{sec:next}

In light of the scope and the pervasiveness of the problem of grammar,
nothing short of a series of radical changes in the basic assumptions,
the theoretical stance, and the empirical methodology of linguistics
will suffice to set things straight. In this section, I sketch an
alternative view of language, motivated by the above diagnosis, and
outline a four-strand research plan that aims to elucidate the
psychology of language, its computational basis, its evolution, and
its brain mechanisms.

\subsection{How radical?}

How radical should our collective conceptual change of heart be?  Some
writers in formalist linguistics \citep[e.g.,][]{Phillips10} do
consider alternative explanations for syntax (usually ones that are
rooted in ``performance'' constraints). Others, who are particularly
impressed by psycholinguistic findings on language use, relegate
grammar to a role that is secondary to pragmatics \citep{HornWard04},
as in the ``syntax last'' hypothesis \citep{BeverSanzTownsend98} or in
the idea of ``opportunistic processing'' \citep{Jackendoff11,Acuna16}.

On a more fundamental level, there are linguists who call for a
reevaluation of the basic methodology of linguistics, while retaining
the concept of grammar
\citep[e.g.,][]{Postal04,Haider16}.\footnote{Albeit perhaps a
  model-theoretic, or descriptive, rather than a generative one
  \citep{Postal08}.} Approaches that are more strongly revisionist
reject certain key formalist (Minimalist) tenets, such as the role of
recursion, while retaining the core dogma: according to
\citet[p.54]{Everett17}, grammar is still there, as ``a set of
organizing principles for constructing complex utterances out of
symbols.''

My feeling is that even such relatively far-reaching measures will not
do: a more radical change is needed. Rejecting the grammar dogma and
its corollaries does have some precedents in the science of
language. For instance, \citet{Sampson07} writes, ``I believe that the
concept of `ungrammatical' or `ill-formed' word-sequences is a
delusion, based on a false conception of the kind of thing a human
language is.'' A better conception of grammar and of communication
that it ultimately serves is offered by \citet{LaPolla15}, who views
communication ``not as coding and decoding, but as ostension and
inference.''\footnote{``That is, one person doing something to show
  the intention to communicate, and then another person using
  abductive inference to infer the reason for the person's ostensive
  act, creating a context of interpretation in which the
  communicator's ostensive act ``makes sense'', and thereby inferring
  the communicative and informative intention of the person''
  \citep[p.31]{LaPolla15}.} This process is dynamic and distributed
across the participants in a conversation
(\citealp{RamscarBaayen13,Linell13,Linell16}; this is a key
observation, which underlies also Du Bois' \citeyearpar{DuBois14}
notion of ``dialogic'' syntax).

An entirely unorthodox conception of language, which incorporates many
of the ideas just mentioned, but also decidedly rejects the grammar
dogma, motivates the ``integrational'' linguistics of Roy Harris, who
wrote that ``From an integrational perspective, there are no `rules of
grammar'.''\footnote{Quoted from
  \url{http://www.royharrisonline.com/integrational_linguistics/integrationism_introduction.html}. For
  a sample of the body of work produced within the integrational
  framework, see the special issue of Language Sciences for July 2011
  (vol.~33(4):475-724).} However, empirical work within the
integrationist framework is typically descriptive rather than
explanatory, while some of its theoretical strands
\citep[e.g.,][]{Mendoza11} amount to a return to Skinner's
behaviorism.

The framework that I outline next takes a similar stance on grammar,
pragmatics, ostension and inference, and dynamic interaction among
speakers/listeners, while insisting on the need for a computational
account for the postulated synthesis. 

\subsection{An alternative conception of language}
\label{sec:alt}

I introduce the proposed framework via answers to the series of
questions posed in the beginning of this paper.

\begin{itemize}
  
\item \textbf{What language is for.} Language has emerged, and
  continues to evolve, under evolutionary pressure to serve as a set
  of tools for dynamically and occasionally strategically influencing
  the behavior of others --- that is, for communication, construed not
  as the exchange of coded messages but as intervening in the others'
  thought processes --- and of self.\footnote{Interestingly, even
    verbal reasoning seems to fall under this rubric
    \citep{MercierSperber11}.}  A related view is that

  \begin{quote}
    Language works by presenting and manipulating cultural affordances
    that will cause one's dialog partner(s) to see and do what the
    speaker intends to be seen and done \citep{Anderson16}.
  \end{quote}

\item \textbf{What language is like.} Language is situated,
  incremental, dynamically constrained, concurrent, multimodal social
  behavior that is subject to interpretation:

  \begin{quote}
    A language is a second-order cultural construct, perpetually
    open-ended and incomplete, arising out of the first-order activity
    of making and interpreting linguistic signs, which in turn is a
    real-time, contextually determined process of investing behaviour
    or the products of behaviour (vocal, gestural or other) with
    semiotic significance \citep[p.530]{Love04}.
  \end{quote}

  As any other behavior, it is both structured and constrained in its
  structure, albeit to a greater extent than in other task domains.
  Importantly, its structure is ``open-ended and incomplete'' --- a
  key characteristic, which alone rules out the grammar dogma. Compare
  \citep[p.10]{Sampson07}: 

  \begin{quote}
    [\dots] The grammatical possibilities of a language are like a
    network of paths in open grassland. There are a number of heavily
    used, wide and well-beaten tracks. Other, less popular routes are
    narrower, and the variation extends smoothly down to routes used
    only very occasionally, which are barely distinguishable furrows
    or, if they are used rarely enough, perhaps not even visible as
    permanent marks in the grass; but there are no fences anywhere
    preventing any particular route being used, and there is no sharp
    discontinuity akin to the contrast between metalled roads and
    foot-made paths.
  \end{quote}

  I shall return to discuss this metaphor later.
  
\item \textbf{What it means to know language.} To know language is to
  have a set of skills for intentionally influencing behavior and
  thought.\footnote{Clearly, this requires that some prior knowledge
    be shared among the users of language. The form that such
    knowledge may take, the computational processes that make use of
    it, and their possible brain basis are discussed in
    \citep{Edelman17monster}.}

\end{itemize}

\subsection{A work plan}

I shall now outline the four key components of a research program
intended to substantiate the proffered working hypotheses about
language. The four themes are the ecological and ethological
characteristics of language, the computational processes that support
it, the evolutionary roots of language, and its brain basis.

\subsubsection{Ecology and ethology of language}


Although explaining language will require coordinated understanding on
all four fronts, priority may be given to the study of linguistic
behavior, because it can serve as a rich source of constraints on
computational theory and neurobiology. The new psycholinguistics
should recognize the multiple factors that shape language, including
human nature, culture, and the external environment
\citep{Steffensen14} and develop a methodology that would do justice
to the socially situated, multidimensional data generated by language
use in realistic settings \citep[p.546]{Weigand11}. Both these
challenges are not yet widely recognized, let alone addressed.

The notion that language should be studied as a dialogic, social
phenomenon (which requires dropping the pretense that isolated
sentences or monologues will do) is gaining recognition in
psycholinguistics. Publications in this area are still mostly
programmatic
\citep[e.g.,][]{PickeringGarrod04,Linell13,Steffensen14,DuBois14}, in
part because not much realistic dialogue-derived data are available (a
prominent exception is Du Bois's Santa Barbara Corpus).\footnote{While
  \citet{PickeringGarrod04} call their proposal a ``model,'' their
  coverage of the architecture of the language system (section 7.2)
  consists of three paragraphs, which are devoted entirely to an
  argument in favor of Jackendoff's \citeyearpar[fig.6]{Jackendoff11}
  ``parallel architecture'' and against Chomskian
  ``syntacticocentrism'' (Jackendoff's term).} Typically, these works
describe theoretical frameworks centered around rather abstract
principles and concepts such as interactive alignment in production
and comprehension \citep{PickeringGarrod04}, the ``non-locality''
\citep{Steffensen15} and the distributed nature of ``languaging''
\citep{Linell13}, or even just the dialogicity of syntax
\citep{DuBois14}.

Both the collection and the use of empirical data in the
dialogic/distributed approaches to language face unique difficulties.
First, naturalistic language behavior is necessarily multidimensional:
conversations involve multiple, parallel, asynchronously unfolding
strands of information \citep{KolodnyEdelman15} and may be conducted
in complex ecological (physical and social) settings, which all lead
to a flood of data. Second, conversations often depend on the
speakers' shared and/or unique personal history \citep{Steffensen15},
which makes a corpus that neglects to somehow take it into account
necessarily incomplete. Third, the overwhelming combinatorics of all
these factors, which conspire to make every conversation unique,
necessitates supplementing the popular ``big data'' approaches in
empirical science and engineering with ones that work for particulars
or isolated case studies, an approach that \citet{Steffensen16} has
dubbed ``cognitive probatonics.''\footnote{From the Greek
  \textit{pr{\'o}bat{\'o}n}, ``a single sheep'' (Luke 15,4--6).}

\subsubsection{Computational modeling}

Having discarded the notions of syntactic well-formedness and ``strong
generativity'' along with the grammar dogma, we are not obligated to
seek models that would reproduce precisely some hypothetical hidden
structure of the utterances encountered in what will be necessarily
extremely complex corpus data. What alternative theoretical
considerations should then guide computational modeling and what
should be its goal?

Because in the brain language comprehension seems to be intimately
interrelated with production \citep{HassonEtAl14}, a good model should
cover both. A possible goal here is ``weak generativity'': modeling
the production ``merely'' of the surface form of utterances,
preferably ranked by their probabilities relative to some reference or
training corpus \citep{WaterfallEtAl10}. According to this approach,
which has been called the ``new Empiricism'' \citep{Goldsmith07emp}, a
weakly generative model is evaluated on its perplexity, or the
reciprocal average per-word probability that it assigns to sentences
in a withheld test corpus.

On a closer look, however, this formulation is seen to be as detached
from the reality of naturalistic language use as the formalist
fictions. A weakly generative modeling framework typically takes the
form of a probabilistic grammar, uses sentences as units of
evaluation, and sneaks in through the back door (a graded version of)
well-formedness by pretending that the probabilities of sentences are
fixed over time and independent of context.


Some recent developments in computational linguistics attempt to bring
language modeling a little closer to reality. One way of doing that is
to model the pragmatics of conversation, for instance by endowing the
model with a set of beliefs about the state of affairs being discussed
(as in game theory), the beliefs being modified after each exchange
according to the Bayes rule \citep{GoodmanFrank16}. This formalism
may, however, be too restrictive: as \citet[p.23]{Sampson07} points
out, ``a natural language does not embody fixed inference
relationships between its statements, such that from a given set of N
natural-language premisses (whether N is zero or a larger number)
there is some definite set of conclusions which can be validly drawn
from them.''

An antidote to such over-formalization is to try to keep logic out of
it as far as possible. Thus, \citet{PapernoEtAl16} observe that
``[word prediction] is a much simpler setup than the one required,
say, to determine whether two sentences entail each other'' and
propose an evaluation task in which the model is given a paragraph or
so of context and a target sentence from which one word is omitted;
the goal is to guess the omitted word. State of the art models are
``effectively unable'' to solve this task, yielding a $0$ percent
correct performance. 


The realization that slavishly generating ``likely'' sequences of
words is not a useful approach to modeling language requires a
reevaluation of Lashley's \citeyearpar{Lashley51} celebrated notion of
the problem of serial order in behavior. Traditional solutions to the
problem as it has been originally formulated can be classified as
associative (based on response chaining between successive elements),
positional (based on coding the absolute order of elements in the
sequence), and hierarchical (based on multiple levels of abstraction
in representing sequential order), Lashley himself favoring the
latter. Given that linguistic behavior is multimodal, socially
situated, and dynamically controlled, these solutions are all
fundamentally inadequate \citep{KolodnyEdelman15}.

A rather different approach to modeling language emerges when the
\emph{problem} of serial order is recast in light of these
observations. This approach, competitive queuing (CQ), aims to show
that serial order can arise out of a parallel representation through a
series of winner-take-all steps, in each of which the next element in
the would-be sequence is selected via competition
\citep{HoughtonHartley96,Bullock04}. Notably, the basic elements can
be quite sizeable themselves, as in formulaic (set) expressions or
partially lexicalized constructions, which are known to play a key
role in natural language learning and use
\citep{Gross94,Tomasello03book}.

Different model architectures may be particularly good at addressing
distinct aspects of the control problem that underlies language
behavior. For instance, the simple two-layer competitive queuing (CQ)
network \citep[fig.1]{BohlandBullockGuenther10} is effective in
integrating multiple constraints that join forces in generating a
sequence of actions from a parallel representation, while recurrent
(folded) networks offer representational savings, and transition
(unfolded) networks are easier to learn and control. These
considerations motivate a hybrid, division-of-labor approach to
modeling language. 

One such hybrid approach is exemplified by the model of speech
production of \citep{BohlandBullockGuenther10}, which reaches up to
the level of the phonology of isolated words, stopping short of
sentence construction.  This model, which is capable of generating
hierarchically and contextually constrained sequences of categorical
elements of speech, integrates CQ into a transition network
architecture.  \citet[p.1522]{BohlandBullockGuenther10} note
explicitly that their model ``combines elements of both CQ and
positional models,'' as well as ```serial chain' representations.''
The model is not learned; instead, the network weights that determine
its dynamics are ``hand-wired'' --- manually set to values that
guarantee the desired outcomes when the differential equations
describing the network are integrated
\citep[p.1516]{BohlandBullockGuenther10}.

The second example is a model of machine translation developed and
tested by \citet{EdelmanSolan09}. This model, which incorporates the
ADIOS algorithm \citep{SolanEtAl05pnas}, learns structured transition
networks for the source and target languages, then uses the activation
pattern induced by the input utterance in the source network to
selectively activate elements (constructions) in the target
network. The resulting spread of activation in the target network is
subject to the patterns of usage in the target language. This process
generates a probability-ranked list of competing outputs, which is
then reranked so as to take into account any additional thematic or
contextual constraints. As a result, this model produces utterances
that are both structurally tractable and semantically apt, insofar as
they combine familiar constructions with abstract meaning (captured by
the many-to-many source-to-target concept map) and on-the-fly
context-specific control.

\subsubsection{Evolution of language}

As a natural biological phenomenon, language can only be understood in
the light of evolution \citep{Dobzhansky73}. In the absence of a
fossil record such as corpora that predate writing,\footnote{Note that
  in any case written language differs from spoken language in
  register and much of its structure \citep{Sampson07}.}  theorizing
about language evolution has long been considered futile.  On the kind
of integral account propounded here, this task would seem to be even
more hopeless: even a real, naturalistic corpus of language may be of
little use, because the original import of utterances that comprise it
is lost when taken out of situational context and disconnected from
the shared knowledge and individual experiential histories of the
participants.

Despite the scarcity of data, a lively debate has been under way for a
while now about the nature of language evolution and its relationship
to the evolution of other behaviors and of the brain \citep[for a
  sample of the literature,
  see][]{PinkerBloom90,Lieberman02b,HauserChomskyFitch02,ChristiansenKirby03,JackendoffPinker05,FitchHauserChomsky05,MerkerOkanoya07,ChristiansenChater08,Tallerman14,BolhuisEtAl14}.
Progress on this front depends on conceptual ``triangulation'':
bringing to bear on each other all conceivable sources of constraints
regarding the nature and interaction of potentially relevant traits,
along with the task contexts in which these traits could have been
applicable. Unlike the formalist postulate of a point mutation (which
supposedly created the Merge operation and thereby immediately made
recursive grammar possible; \citealp{BolhuisEtAl14}), which exists in
a conceptual vacuum, the idea of a multiply constrained combination of
reusable traits and of gradual build-up of language-related skills is
both more plausible and more promising methodologically.

One of the directions that this latter approach can pursue has to do
with the multimodal nature of language \citep{KolodnyEdelman15}, which
is only reducible to sequences of symbols when confined to a dead
corpus. Thus, \citet{DaleEtAl16} propose ``what may be a crucial
ingredient in language evolution: multimodal synergy.'' Another
potentially rich source of constraints is the social context in
language operates
\citep{Cangelosi01,CataniBambini14,KirbyEtAl15}. Furthermore, language
acquisition and evolution can and should be addressed together
\citep{LockeBogin06,ChaterChristiansen10,Pinker10}, along with the
role of the brain in shaping language
\citep{ChristiansenChater08}. Finally, a careful analysis of the
social dynamics of language use may explain the ``major transition''
\citep[p.231]{SzathmaryMaynardSmith95} in human behavior from costly
to conventional signaling by increasing reliance on social enforcement
of honesty and by the relegation of meanings from unitary to
combinatorial/composite signals \citep[p.13193]{LachmannEtAl01}.

\subsubsection{The brain mechanisms of language}

To really understand the place of language in the brain, and thereby
extend the science of language to include the implementation level of
the Mayr/Tinbergen/Marr explanatory framework, we must do more than
list the putative brain areas involved and offer verbal descriptions
of their respective roles (as, for instance, when mere identification
of lexical memory and sequential processing with different brain areas
is called a ``model''; \citealp{Ullman01}).  The model must be
explicit in spelling out the linking principles \citep{PoeppelEtAl12}
between behavioral data and neural computation and must map the
posited computations, not vaguely, to areas \citep[as
  in][]{Ullman01,Hagoort14,Fedorenko14}, but specifically, to brain
\emph{circuit-and-synapse} diagrams \citep[p.262]{KolodnyEdelman15}.

When viable models start to emerge, they will transcend the textbook
corticocentric view of the brain basis of language
\citep[e.g.,][]{BornkesselSchlesewsky13,LelyPinker14}, with its
exclusive focus on the Broca and Wernicke areas. The narrow focus on
the cortex in language is no longer tenable \citep{Lieberman02},
especially in the light of new data \citep[e.g.,][]{HassonEtAl14}
documenting the extensive spread of language production and
comprehension over the brain.\footnote{Nor is corticocentrism a
  tenable approach to any other behavior \citep{Edelman15}.}
Furthermore, on the emerging understanding of the ubiquity of neural
reuse in brain function \citep{Anderson10}, the question that any
model should address is not where in the brain is the ``language
center,'' but rather how the major circuits implicated in other
behaviors contribute to language. All these circuits involve (in
addition to premotor and motor cortical areas) other brain structures,
notably the basal ganglia, the hippocampus, and the thalamus.


\paragraph{Basal ganglia.} Among the most prominent circuits 
involving the basal ganglia is the set of loops that connect the
striatum, globus pallidus, substantia nigra, and subthalamic nucleus
to the thalamus, all of the cortex, and back to the striatum
\citep{Wickens97}. This circuit is anatomically intricately
structured: there are four major subdivisions \citep{AlexanderEtAl86},
which are further differentiated into many ``stripes''
\citep{OReillyFrank06,Botvinick08}. Their functions in composing
behavior include the inhibition of competing alternatives
\citep[e.g.,][]{Botvinick08}, the sequencing of elementary actions
\citep[e.g.,][]{AldridgeBerridge98,NakaharaEtAl01,AldridgeEtAl03}, and
parsing and concatenation of subsequences \citep{JinEtAl14}. All these
functions are central to language generation and processing, as, for
instance, when a choice must be made from among several constructions
(a likely role of the striatum in the language system;
\citealp[cf.][]{TylerEtAl05}) or a construction is abandoned in favor
of another (which may be supported by `` split circuits'' in the
cortical-basal-thalamic-cortical loops;
\citealp[Fig.5]{JoelWeiner94}).\footnote{A useful mental image for
  such switching, usually branded as ``disfluency,'' is the mining
  cart chase scene in the film \textit{Indiana Jones and the Temple of
    Doom} \citep[p.301]{Edelman08book}. On this view, constraints on
  language production are more like those on traffic in a railroad
  yard or a subway system \citep[p.274]{Edelman08book} than on the
  traversal of the ``open grassland'' of \citep[p.10]{Sampson07},
  mentioned earlier.}  Moreover, these functions are exactly what it
takes to implement competitive queuing
\citep{BullockEtAl09,BohlandBullockGuenther10}.


\paragraph{Hippocampus.} The classical view of one function of the
hippocampus having to do with memory is certainly compatible with its
documented role in lexical and referential aspects of language
\citep{BreitensteinEtAL05,KurczekEtAl13}. Since at least
\citep{Levy96}, the notion has been entertained that the other
function classically attributed to the hippocampus, navigation, can be
of critical importance to sequence learning and replay
\citep[e.g.,][]{AgsterEtAl02,FortinAgsterEichenbaum02,RodriguezEtAl04,GheysenEtAl10,DeVitoEichenbaum11,AlbouyEtAl13},
and therefore to language, which the hippocampus orchestrates in
coordination with the striatum \citep{AlbouyEtAl08,AlbouyEtAl13}.
Neurological case studies show that hippocampus lesions in early
childhood interfere with language acquisition
\citep{DelongHeinz97,WeberEtAl06}. Some of the plausible
language-related specializations of the hippocampus are construction
learning \citep{SegerCincotta06} and their flexible use in generation
\citep{DuffBrownSchmidt12,PezzuloEtAl14}, especially sequence
composition and hierarchical organization and control (by
extrapolation from rodent data;
\citealp{WilsonEtAL09,GuptaEtAl10,DezfouliBalleine13}; see
\citealp{AllenEtAl14}).  Not only there is a functional link between
navigation and language (as indicated by the verbal shadowing task
interfering with wayfinding; \citealp{MeilingerEtAl08}; \citealp[see
  also][]{Arsenjevic08}): the hippocampus also supports navigation in
abstract spaces, as in sequential olfactory decision task of
\citep{AgsterEtAl02}; see \citep[p.760]{FriedEtAl97} for an early
insight and \citep{DabaghianEtAl14} for a recent synthesis. Finally,
the autoassociative dynamics of some hippocampal circuits
\citep{PfeifferFoster15} is compatible with the use of multiscale
dynamics for organizing sequential behavior
\citep{HuysPerdikisJirsa14}, in a manner that resembles competitive
queuing.


\paragraph{Thalamus.} No cortical function is, or can be,
carried out without a deep involvement of the thalamus
\citep{ShermanGuillery02,Sherman12}. Being an integral part of the
cortico-basal-ganglia loops \citep[fig.2]{HaberCalzavara09}, the
thalamus is indispensable for the shaping of sequential behavior,
including language
\citep{Grosson13,HebbOjemann13,KlostermannEtAl13,BarbasEtAl13};
\citep[see also][p.763]{BullockEtAl09}.

\paragraph{Premotor and motor areas.}
While the returning projections from the thalamus reach all of the
cortex, the cortical areas commonly termed motor play a special role
in the structuring of language, over and above its physical
articulation and timing \citep{MendozaMerchant14}. In particular, the
neural mechanisms in premotor and supplementary motor areas
\citep{Miyashita03,TanjiShima94} appear well-suited for such
language-related functions as chunking (needed for learning
constructions) and sequential planning
\citep[p.127]{Graybiel98}. Because sequential learning in the basal
ganglia is slow \citep[p.131]{Graybiel98}, sequences may first be
captured by the supplementary motor area (SMA;
\citealp{TanjiShima94}), then consolidated in the basal ganglia.

\section{Concluding remarks}


Two of the phrases that I drafted to serve in the title of the present
paper --- \textit{Verbal Behavior} (as in Skinner, 1957) and
\textit{Syntactic Structures} (as in Chomsky, 1957) --- stand for two
big ideas, whose clash happened so long ago that the victory of the
latter and its subsequent turning into an orthodoxy are rarely
questioned anymore. This is too bad: while most of Chomsky's
\citeyearpar{Chomsky59} arguments against Skinner stand, the
alternative that he promoted turned out to be fatally flawed.  Its
failure stems from its being founded on the dogma of grammar; and the
occasional rebellions against it fall short because they fail to
completely reject the dogma.

In trying to formulate a real alternative, we must not be discouraged
by the overwhelming complexity of language that becomes apparent once
the virtual reality show generated by the grammar dogma is shut
down. It does seem that language, not consciousness, is \emph{the}
``hard problem''\footnote{In consciousness research, this problem ---
  explaining phenomenality or qualia \citep{Chalmers95hp} --- may not
  be hard at all, if it is indeed boils down to a simply statable
  informational \citep{OizumiAlbantakisTononi14} or
  dynamical/geometric \citep{FeketeEdelman11} principle.} of the
cognitive and brain sciences, but at least as long as we attack the
real problem and not an imaginary one, we may attain real progress.

\subsubsection*{Acknowledgments}

I thank Barbara Finlay and Oren Kolodny for their comments on some of
the views stated and defended here and Christina Behme for helpful
editorial remarks. The ideas developed in this paper were first
presented at the 2nd Conference on Usage-Based Linguistics, Tel Aviv
University, June 2016. The poem that serves as the epigraph
accompanies Warren McCulloch's \citeyearpar{McCulloch65freud} remarks
on Freud (\textit{The Past of a Delusion}), whose impact on psychology
and the humanities is often compared to that of Chomsky.

\theendnotes

\vskip 0.3in
\bibliographystyle{chicago}
\bibliography{/Users/shimon/Documents/my}

\end{document}